\documentclass[runningheads]{llncs}
\usepackage[T1]{fontenc}
\usepackage{graphicx}
\usepackage{amssymb}
\usepackage{booktabs}
\usepackage{array}     
\usepackage{tabularx}  
\usepackage{float}
\usepackage{marvosym}
\usepackage{cite}
\usepackage{xcolor} 

\usepackage[
    colorlinks=true,
    linkcolor=blue,
    citecolor=blue,
    urlcolor=black,
    bookmarks=true,
    bookmarksnumbered=true
]{hyperref}
\hypersetup{bookmarksdepth=subsection}

\newcolumntype{Y}{>{\centering\arraybackslash}X} 

\begin{document}

\title{HiGS: Hierarchical Generative Scene Framework for Multi-Step Associative Semantic Spatial Composition}
\titlerunning{HiGS}

\author{Jiacheng Hong\textsuperscript{(\Letter)} \and
Kunzhen Wu \and
Mingrui Yu \and
Yichao Gu \and
Shengze Xue \and
Shuangjiu Xiao\textsuperscript{(\Letter)} \and
Deli Dong\textsuperscript{(\Letter)}}

\authorrunning{J. Hong et al.}

\institute{Digital ART Lab, Shanghai Jiao Tong University, Shanghai, China\\
\email{\{rakutenipoi,xiaosj,arli\}@sjtu.edu.cn}\\
\url{http://dalab.se.sjtu.edu.cn/www/home/}}
\maketitle              
\begin{abstract}
Three-dimensional scene generation holds significant potential in gaming, film, and virtual reality. However, most existing methods adopt a single-step generation process, making it difficult to balance scene complexity with minimal user input. Inspired by the human cognitive process in scene modeling, which progresses from global to local, focuses on key elements, and completes the scene through semantic association, we propose HiGS, a hierarchical generative framework for multi-step associative semantic spatial composition. HiGS enables users to iteratively expand scenes by selecting key semantic objects, offering fine-grained control over regions of interest while the model completes peripheral areas automatically. To support structured and coherent generation, we introduce the Progressive Hierarchical Spatial-Semantic Graph (PHiSSG), which dynamically organizes spatial relationships and semantic dependencies across the evolving scene structure. PHiSSG ensures spatial and geometric consistency throughout the generation process by maintaining a one-to-one mapping between graph nodes and generated objects and supporting recursive layout optimization. Experiments demonstrate that HiGS outperforms single-stage methods in layout plausibility, style consistency, and user preference, offering a controllable and extensible paradigm for efficient 3D scene construction.

\keywords{Hierarchical Scene Generation  \and Associative Semantic Spatial Composition \and Progressive Graph-Guided Generation \and Interactive Incremental Editing}
\end{abstract}

\section{Introduction}
The design and generation of three-dimensional scenes has long been a complex process that heavily relies on human expertise, with significant applications in gaming, film production, and virtual reality. High-quality 3D scenes must not only achieve visual realism but also maintain semantic plausibility in their spatial structure. The scene layout, object scales, and support and contact relationships between objects must all adhere to physical and semantic constraints to ensure the scene’s usability and interactivity. Traditional 3D scene modeling relies on manual design, requiring designers to possess strong spatial construction skills and extensive domain knowledge. This makes the modeling process time-consuming and labor-intensive, while limiting the ability to quickly iterate or generate diverse scenes. Procedural generation methods can alleviate some of this workload, but their generality and flexibility remain limited, making it challenging to produce scenes that are both structurally reasonable and stylistically diverse according to specific task requirements or user preferences. 

In recent years, the rapid development of deep learning and generative models has driven research in automated 3D content generation. Many methods\cite{para2023cofs,paschalidou2021atiss} explore end-to-end 3D scene generation strategies, learning scene layout patterns directly from data. Some methods\cite{fridman2023scenescape,huang2025dreamfuse,hollein2023text2room,shriram2024realmdreamer,yu2025wonderworld,yu2024wonderjourney} applies diffusion models\cite{rombach2022high} to 3D scene generation tasks, benefiting from the diffusion model’s ability to encode diverse scene structures and significantly improving visual detail and style variety. However, dense indoor or scene-level generation still faces challenges such as occlusion, object completeness and semantic coherence.

Meanwhile, large language models (LLMs) have demonstrated strong capabilities in text understanding and knowledge reasoning, making “text-driven scene generation” possible\cite{feng2023layoutgpt,yang2024holodeck,zhou2025roomcraft,zhou2024gala3d,ocal2024sceneteller}. Nonetheless, LLMs still exhibit limitations when handling tasks involving complex spatial structures, particularly due to the lack of explicit modeling of 3D spatial relationships. To compensate for LLMs’ shortcomings in spatial understanding, vision-language models (VLMs) have been introduced into 3D scene generation\cite{ling2025scenethesis,yang2024scenecraft,sun2025layoutvlm}. VLMs integrate visual evidence with textual prompts, providing technical support for object recognition, localization, and mask-level segmentation, and offering visual grounding for text-based spatial reasoning, thereby enhancing the model’s understanding and control of spatial layouts.

Despite these advances, existing 3D scene generation methods still face several key challenges: 
\begin{itemize}
  \item \textbf{High demands on user input}: Human scene modeling naturally follows a coarse-to-fine, layered process of iterative refinement. Single-step generation forces users to specify all details upfront, increasing the input burden and misaligning with natural workflows.
  \item \textbf{Limited controllability and interactivity}: Single-step generation provides limited support for incremental or fine-grained local adjustments. Users often focus on specific areas and expect non-critical regions to be completed automatically, but current methods lack effective local editing and adaptive generation, which constrains overall controllability.
  \item \textbf{Neglect of hierarchical scene structure and progressive semantics}: Mainstream approaches usually generate entire scenes in a single step, without modeling or optimizing from global layouts to local details, which can lead to structural inconsistencies and uncoordinated details.
  \item \textbf{Lack of dynamic scene understanding and feedback mechanisms}: Existing methods treat the scene images generated by diffusion models\cite{ho2020denoising,rombach2022high} as static inputs, lacking dynamic perception and adaptive correction mechanisms for layout plausibility during 3D scene generation, which limits their effectiveness in handling complex and variable scene requirements.
\end{itemize}

To address these limitations in expressiveness and controllability, we propose HiGS, a hierarchical generative scene framework for multi-step associative semantic spatial composition. Unlike mainstream single-step generation strategies, HiGS progressively enriches scene content across multiple scales through semantic progression and associative reasoning, performing stepwise generation and evaluation from global layouts to local details. This approach achieves a better balance among controllability, editability, and semantic consistency. The progressive, associative workflow substantially reduces the initial input burden on users, enabling them to start with concise functional requirements, iteratively refine preferences based on intermediate results, and rely on the model to automatically complete non-focused regions.
The main contributions of this work can be summarized as follows:
\begin{itemize}
    \item We propose \textbf{a hierarchical 3D scene generation framework} based on multi-step associative semantic spatial composition, supporting progressive refinement from global structure to local details and flexible editing of user-focused regions, enhancing controllability, scalability, and interactivity.
    \item We design \textbf{a progressive hierarchical spatial-semantic graph} to organize scene semantics across multiple scales and serve as an intermediate representation for generation and optimization. This graph guides object placement and hierarchical modeling, supports object-level addition, deletion, and modification, and propagates local edits to the global layout.
    \item We introduce \textbf{a multi-scale scene alignment and optimization algorithm} to dynamically detect and adjust the spatial plausibility of newly generated regions during incremental generation, ensuring overall structural coherence, semantic consistency, and layout stability.
\end{itemize}

\section{Related Work}

\subsection{3D Scene Generation}

In the field of 3D scene generation, some approaches employ Transformer models to generate scene layout information\cite{para2023cofs,paschalidou2021atiss}. In recent years, numerous studies have applied diffusion models to indoor scene generation. Among these, DreamFusion\cite{huang2025dreamfuse} introduced diffusion models to 3D generation, while subsequent works\cite{fridman2023scenescape,hollein2023text2room,shriram2024realmdreamer,yu2025wonderworld,yu2024wonderjourney} further incorporated depth information obtained from depth estimators\cite{bae2022irondepth,bhat2023zoedepth,ke2025marigold} as a constraint, improving the quality and realism of generated scenes. These data-driven generative models greatly enhance the richness of scene details and physical plausibility. however, they generally exhibit the following limitations: (1) scenes are primarily generated as holistic visual reconstructions, lacking segmentation and modeling of individual objects, which hinders user interaction and editing. (2) there is little active perception or optimization of scene layouts, making it difficult to dynamically detect and correct spatial inconsistencies.

\subsection{LLM-Based Scene Generation}

With the development of LLMs and VLMs, an increasing number of studies attempt to leverage their strong knowledge and reasoning capabilities to guide 3D scene generation\cite{feng2023layoutgpt,yang2024holodeck,zhou2025roomcraft,zhou2024gala3d,ocal2024sceneteller}. HOLODECK\cite{yang2024scenecraft} and RoomCraft\cite{zhou2025roomcraft} focus on modeling reasonable object arrangements in scenes using LLMs, while GALA3D\cite{zhou2024gala3d} allows users to flexibly modify scene content through their response. While some methods\cite{ling2025scenethesis,sun2025layoutvlm,yang2024scenecraft} extend this approach by incorporating visual models. Scenethesis\cite{ling2025scenethesis} uses spatial priors from image generation models to guide scene generation, and SceneCraft\cite{yang2024scenecraft} introduces bounding-box constraints to enhance spatial expressiveness. These methods generally rely on LLMs/VLMs to extract spatial relationships among objects, optimizing overall layout.

However, they mostly treat LLMs as evaluative tools for existing content, without fully leveraging their rich commonsense knowledge and generative potential. Due to this design, current methods face two main challenges in practice. On one hand, users must provide long and detailed textual descriptions at the initial generation stage to tightly constrain outputs. On the other hand, single-step generation strategies lack mechanisms for iterative refinement or progressive improvement, making it difficult for users to incrementally adjust and optimize scenes once outputs deviate from expectations.

In contrast, the HiGS framework proposed in this work centers on multi-step, associative semantic spatial composition and introduces a hierarchical, progressive generation mechanism. HiGS not only allows users to iteratively expand and refine regions of interest based on high-level semantic cues but also relies on the model to automatically complete non-focused areas, enabling flexible interaction and fine-grained control. By dynamically integrating the current scene state and user input at each generation stage, HiGS significantly enhances the controllability, diversity, and interactivity of 3D scene generation.

\section{Method}

\begin{figure}[htb]
\centering
\includegraphics[width=\textwidth]{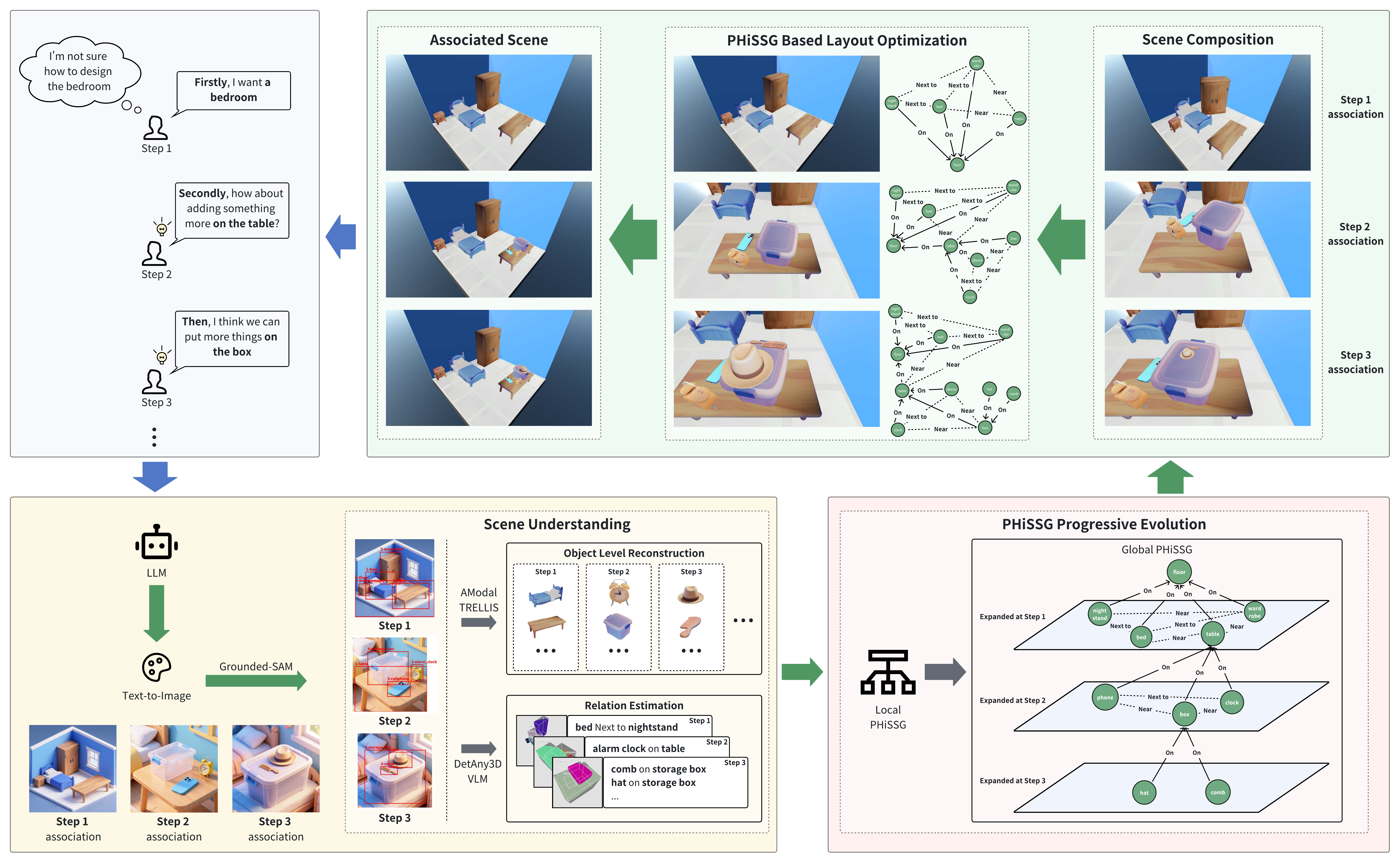}
\caption{Overview of HiGS. } \label{pipeline}
\end{figure}

The HiGS framework generates 3D scenes across multiple semantic hierarchies and levels of detail through a recursive, progressive process. As illustrated in Fig.~\ref{pipeline}, given a scene description text $T_{scene}$, we first parse the text using a large language model and generate an initial scene image via a text-to-image model (Section~\ref{sec:llm-guided-image-generation}). The scene image is then analyzed (Section~\ref{sec:scene-understanding-and-construction}) to extract the geometric structures and physical attributes of individual objects. Based on this analysis, we construct a hierarchical spatial-semantic graph (Section~\ref{sec:progressive-hierarchical-spatial-semantic-graph}) that integrates the scene’s spatial relationships and semantic dependencies, ultimately producing a global 3D scene containing the core semantic elements (Section~\ref{sec:scene-composition}). During the initial generation stage, an isometric view is employed for scene modeling to ensure the global structure’s plausibility and the spatial relationships of major objects.

After the initial global scene is generated, users can freely select regions of interest based on the current layout and specify further refinement requirements. HiGS then re-enters the above analysis and generation process with the user-specified reference objects, progressively completing and integrating local details. This process can iterate over multiple rounds, with each round focusing on newly selected regions of interest, gradually optimizing the scene from global to local and from coarse to fine in a hierarchical manner. It should be noted that, except for the first generation stage, subsequent refinement steps are no longer constrained to an isometric view, but instead adaptively adjust the viewpoint according to the user-focused regions.

\subsection{LLM-Guided Image Generation}\label{sec:llm-guided-image-generation}

HiGS adopts a two-stage LLM strategy to generate 3D scenes from natural language. Given a scene description $T^n_{scene}$, the LLM first produces an object list $O^n=\{o^n_i|o^m_i\in O^n\}$, and then constructs a scene prompt $T^n_{sd}$ based on it. This prompt is used by a diffusion model to generate the scene image $I^n_{scene}$. For the initial scene $S^0$, we apply an additional isometric view constraint $T_{iso}$.

To ensure consistency across generation steps, we introduce a global scene constraint $T^n_{global}$, derived from the current global semantic context. This constraint supplements both LLM stages, guiding the generation of local scenes that align with global style and structure.

The final prompts used to guide scene generation are formulated as:

\begin{equation}
T^0=T_{iso} \cup T^0_{global} \cup T^0_{sd}
\end{equation}

\begin{equation}
T^n=T^n_{global} \cup T^n_{sd} \quad (n > 0)
\end{equation}

\subsection{Scene Understanding and Construction}\label{sec:scene-understanding-and-construction}

The core of scene understanding lies in accurately identifying and modeling the spatial positions and semantic relationships among different objects. We divide this process into two stages: Object Level Reconstruction and Relation Estimation. Through these two stages, we obtain the geometry, spatial orientation, and mutual relationships of each object in the scene, and use them to construct an incremental PHiSSG later, thereby enabling a structured understanding of the scene layout.

\subsubsection{Object Level Reconstruction.}To identify the positions and boundaries of objects from the scene image $I^n_{scene}$, we adopt the Grounded-SAM\cite{ren2024grounded} method for object segmentation. Considering that occlusion is inevitable in real-world scenarios, we use AModal\cite{wu2025amodal3r} to complete occluded or partially missing object regions. After occlusion completion, the completed image is fed into an image-to-3D model TRELLIS\cite{xiang2025structured} to generate a 3D point cloud representation for each target object and to estimate its minimum bounding box, thus obtaining the object’s geometry and size information.

In addition, to extract the spatial position and orientation of each object, we use the DetAny3D\cite{zhang2025detect} model to estimate its spatial center position and yaw angle (rotation around the vertical axis), thereby acquiring the spatial pose of the object in the newly generated local scene.

\subsubsection{Relation Estimation.}After obtaining the spatial poses of all objects, we further identify the spatial-semantic relationships between them. Specifically, we input the current scene image $I^n_{scene}$ together with the geometric and positional information obtained from Object Level Reconstruction into a VLM. The VLM combines visual information with textual priors to infer the spatial relationships among objects. Leveraging these inferred relationships, we construct a local PHiSSG for the current generation step. The detailed methodology for this process will be presented in the following section.

\subsection{Progressive Hierarchical Spatial-Semantic Graph}\label{sec:progressive-hierarchical-spatial-semantic-graph}

To efficiently model and organize structural information in 3D scenes, we introduce the Progressive Hierarchical Spatial-Semantic Graph (PHiSSG), denoted as

\[
PHiSSG=(V,E)
\]

where the vertex set $V$ represents independent entities in the scene $o_i$. Each vertex $v_i\in V$ corresponds to a specific object and contains its geometric attributes and semantic information in 3D space. The edge set $E$ captures spatial and semantic relationships between different objects. Each edge $e_{(i,j)}\in E$ connects two vertices $v_i$ and $v_j$.

For node attributes, each node contains the object’s spatial position $pos\in \mathbb{R}^3$, orientation $rot\in \mathbb{R}^3$, scale $scale \in \mathbb{R}$, category label and a unique node identifier $nid \in \mathbb{Z}$. Edge attributes include the source node (src), destination node (dst), and the specific relation type (relation), enabling precise representation of semantic dependencies and spatial constraints between objects in the graph.

\subsubsection{Spatial Relations.}PHiSSG supports a variety of common geometric and semantic relations, defining \textquotesingle On\textquotesingle\ and \textquotesingle Inside\textquotesingle\ as strong dependency relations. In such relations, the depended-upon object is considered the parent node, and the dependent object is the child node.

At the macro level, strong dependency relations organize the scene into a semantic dependency chain. When the pose of one node changes, other nodes along the dependency chain may also be affected. To maintain physical plausibility and semantic consistency of the spatial structure during multi-step generation, we further design a Layout Optimization mechanism, which performs globally consistent adjustments and constraints along the dependency chains, including recursive pose updates and stability correction.

\subsubsection{Layout Optimization.}To maintain both the physical plausibility and semantic consistency of the scene, we further apply layout optimization strategies that adjust object positions and orientations based on their dependencies and spatial constraints.

\paragraph{Recursive Position Adjustment.}The system records the relative transformation between parent and child nodes. When the parent node’s pose changes, this transformation is directly applied to the child node to restore the relative layout and recursively propagates down the dependency chain until all affected nodes are updated.

\sloppy

\paragraph{Stability Correction.}For parent-child pairs with an \textquotesingle On\textquotesingle\ relation, we introduce geometric and physical stability constraints. Specifically, the geometric center of the placed object must project onto a valid placement area on the supporting surface. In implementation, the system first determines the projected area of the supporting object’s top surface, then calculates the projected position of the geometric center of the placed object on this plane. If the position falls outside the valid area, the system computes the minimal horizontal translation vector to move the geometric center to the nearest valid location, thereby ensuring placement stability and plausibility.

\begin{figure}[htb]
\centering
\includegraphics[width=\textwidth]{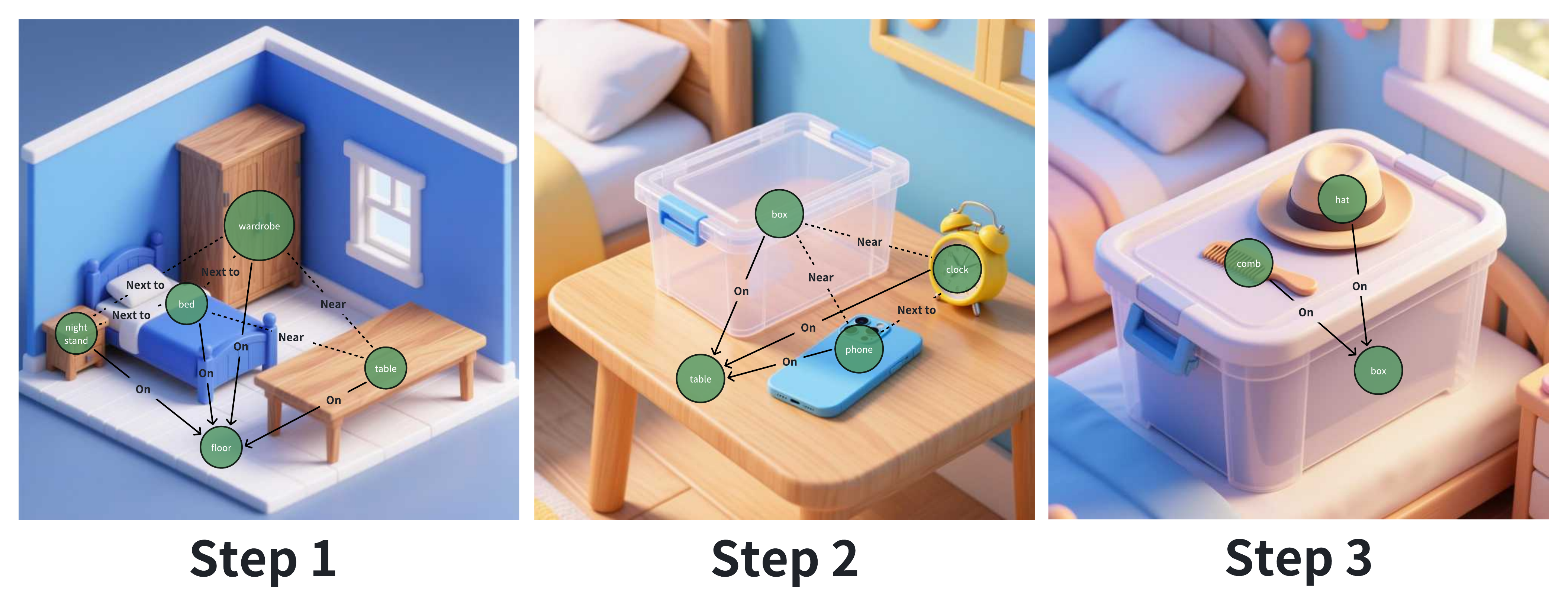}
\caption{\textbf{Local PHiSSG construction over three association steps.} Each step progressively incorporates new objects and updates the graph structure to reflect refined spatial and hierarchical relations. Our method supports an arbitrary number of association steps, enabling flexible and extensible scene expansion.} \label{local_ssg}
\end{figure}

\subsubsection{Progressive Evolution of PHiSSG.}To better support association driven step-by-step modeling and detail refinement in complex scene generation, we propose a progressive evolution strategy, enabling the PHiSSG to evolve dynamically throughout the scene generation process. This strategy organizes the nodes and edges of the graph in a task-driven and adaptive manner.

At the beginning of generation, the system constructs an initial global PHiSSG, where each node represents a key structural object, and edges capture high-level semantic and spatial relationships. At this stage, the PHiSSG serves as an intermediate representation to guide the layout of the 3D scene. When the user selects a specific object as a reference for further modeling, the system generates a finer-grained subscene around the selected region and constructs a local PHiSSG to describe the internal hierarchical structure and spatial dependencies among the newly introduced objects. An example of a local PHiSSG is shown in Figure~\ref{local_ssg}. Layout optimization is then performed within this local graph to ensure semantic consistency and spatial plausibility of the generated region.

Finally, the local PHiSSG is merged into the global PHiSSG. Based on the strong dependency relations, the system determines the hierarchical structure between newly added and existing objects, establishing cross-level semantic and spatial links to form a coherent multi-scale scene representation. The updated global PHiSSG further drives the integration of local content into the global scene and performs global layout optimization, ensuring that local edits are properly absorbed and reflected within the overall semantic structure.

\subsection{Scene Composition}\label{sec:scene-composition}

In the multi-step generation framework, each step produces a local PHiSSG that encodes the geometry and semantics of newly generated objects. The process begins with a user-specified anchor object in the global scene. New objects are generated around this anchor, and a local PHiSSG is constructed to represent their structure and relations. These objects are first placed into a local scene, then integrated into the global context by establishing semantic dependencies with the anchor. The local PHiSSG is merged into the global PHiSSG, and the layout is refined based on the updated graph to ensure overall semantic and spatial consistency.

To support consistent integration, we apply a coordinate alignment strategy that differs across stages. In the initial generation, where the floor is used as the anchor, object orientations inferred from isometric views may exhibit angular deviations from architectural structures. To correct this, we introduce a \textbf{Local Coordinate System Alignment} mechanism based on the assumption that indoor scenes exhibit dominant orthogonal structures. Specifically, we constrain each object's horizontal forward vector $f_i^{xy} \in \mathbb{R}^2$ to a finite set of basis directions:

$$\mathcal{D}=\{\pm d_1,\pm d_2\},\quad where \enspace d_1\bot d_2,\enspace ||d_1||=||d_2||=1$$

Each forward vector is projected onto the ground plane and normalized to obtain $f_i^{xy}$, followed by K-Means clustering (with absolute cosine similarity) to extract dominant direction centers $c_1$ and $c_2$. These are orthogonalized via the Gram–Schmidt process to form $\mathcal{D}$ and each object direction is snapped to its nearest basis using:

\begin{equation}
f_i^{xy}=\arg\max_{d_j\in \mathcal{D}} f_i^{xy}\cdot d_j
\end{equation}

In subsequent steps, where anchor objects are selected from the existing scene, local coordinate alignment is no longer required. For each step, we construct an oriented bounding box for the newly generated region and align it to the anchor object's placement area based on their relative orientation. The corresponding scene structure and semantic relationships, extracted during scene understanding, are naturally integrated into the evolving global PHiSSG.

\section{Experiments}

To evaluate HiGS in multi-scale 3D scene generation, we compare it with GALA3D, a state-of-the-art and publicly available method. GALA3D employs a layout-guided 3D Gaussian representation combined with instance-scene compositional optimization and diffusion priors, enabling it to generate realistic 3D scenes with consistent geometry, texture, scale, and accurate object interactions. In contrast, HiGS leverages the PHiSSG to drive layout adjustment and scene expansion, enabling the generation of structurally sound and semantically coherent 3D scenes without explicit layout input.

\begin{figure}[htb]
\centering
\includegraphics[width=\textwidth]{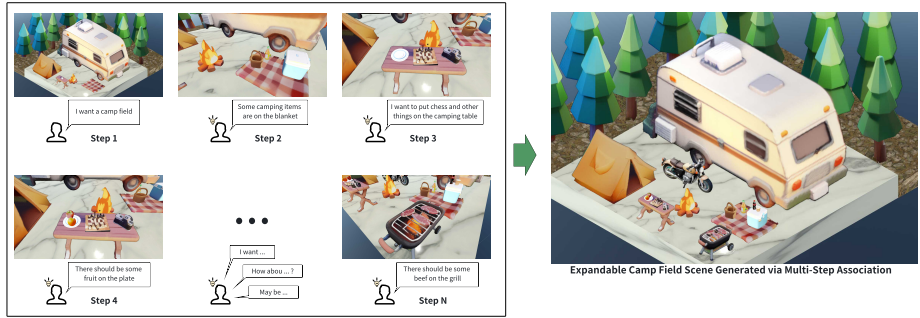}
\caption{Illustrative demo of the associative multi-step generation process with a camp field.} \label{demo}
\end{figure}

A key strength of HiGS is its ability to support unlimited associative expansion, allowing users to iteratively grow the scene around any object. As shown in Figure~\ref{demo}, complex scenes can be dynamically extended in a multi-step manner.

Both methods were tested under identical hardware and runtime conditions, and their generation quality was compared using a combination of subjective questionnaires and multiple objective metrics.

\subsubsection{Implementation.}We use Qwen2.5\cite{qwen} as the LLM, Qwen-Image\cite{wu2025qwenimagetechnicalreport} as the image generation model, and Qwen2.5-VL\cite{Qwen2.5-VL} as the VLM for scene parsing, relation reasoning, and layout-assisted optimization. The scene visualization and user interaction components run in the Unity environment. The experimental hardware platform is an NVIDIA RTX 4090 GPU.

\begin{figure}[htb]
\centering
\includegraphics[width=\textwidth]{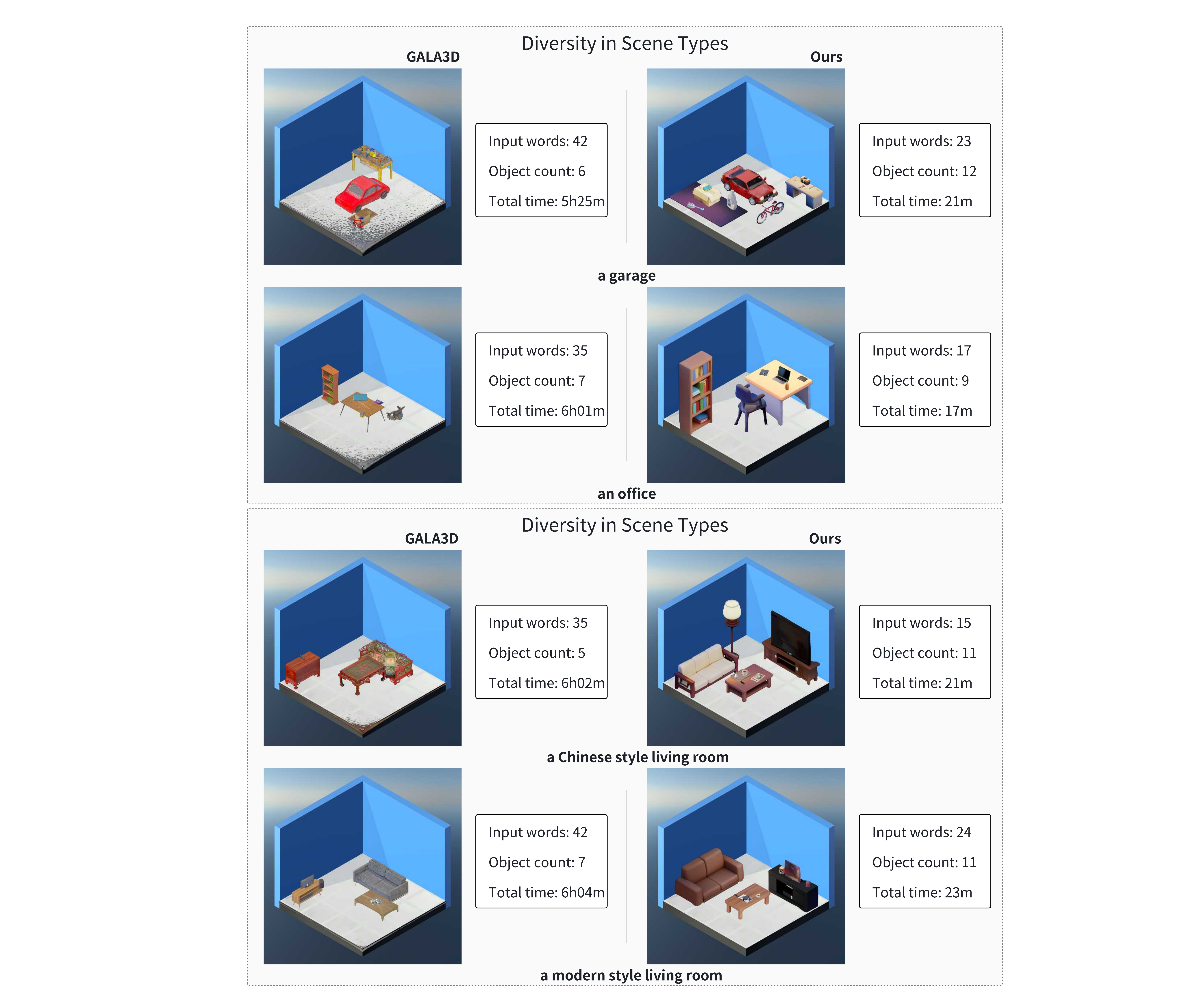}
\caption{Comparative Results of HiGS and GALA3D on Scene Type and Scene Style Diversity.} \label{compare}
\end{figure}

\subsection{Qualitative Comparison}

To systematically evaluate the generation quality and user experience of HiGS, we designed a two-stage subjective user study. In the first stage, participants were presented with 3D scenes generated by HiGS and a baseline method side by side. They were asked to select the one with better visual quality, focusing solely on the objects in the scene while disregarding background factors such as walls, floors, lighting, and shadows. In the second stage, participants conducted a more fine-grained evaluation based on three criteria: layout plausibility (whether the spatial arrangement of objects aligns with real-world logic and whether space is used efficiently), style consistency and aesthetics (whether object styles, colors, and materials are coherent), and complexity (in terms of object count, level of detail, and scene richness). Again, background elements were to be ignored during the evaluation.

We generated 5 distinct 3D scenes and invited 105 participants to complete the evaluation. According to the collected responses, HiGS consistently outperformed GALA3D in terms of layout plausibility, style consistency, and visual appeal, receiving higher ratings from the majority of participants, as summarized in Table~\ref{qualitative_comparison}.

As illustrated in Fig.~\ref{compare}, we compare the generation results of HiGS and GALA3D across various scene types and styles. HiGS appears to perform better in both scene layout and visual consistency. Moreover, HiGS is capable of generating rich and diverse content across multiple hierarchical levels, a feature that GALA3D fails to achieve.

\begin{table}
\centering
\caption{\textbf{Qualitative Comparison.} Scores are in the range of 1 (lowest) to 5 (highest). }\label{qualitative_comparison}
\begin{tabularx}{\textwidth}{lYYYY}
\hline
Method &  Preference$\uparrow$ & Layout$\uparrow$ & Style$\uparrow$ & Complexity$\uparrow$ \\
\hline
GALA3D & 2.86 & 2.60 & 2.78 & 2.69 \\
\textbf{Ours} & \textbf{3.70} & \textbf{3.61} & \textbf{3.76} & \textbf{3.64} \\
\hline
\end{tabularx}
\end{table}

\subsection{Quantitative Analysis}

For quantitative evaluation, we use the CLIP\cite{radford2021learning}, HQ-CLIP\cite{wei2025hq} and BLIP-2\cite{li2023blip} score to measure the semantic consistency between the generated scene and the input text description, with higher scores indicating closer alignment to the input semantics. As shown in Table~\ref{quantitative_analysis}, HiGS outperforms GALA3D across all evaluation metrics, achieving higher CLIP, HQ-CLIP, and BLIP-2 scores, which indicates better semantic alignment between the generated scenes and the input text. Additionally, HiGS generates scenes with a significantly higher average object count, demonstrating its capability to produce more complex and diverse environments.

\begin{table}
\centering
\caption{\textbf{Qualitative Comparison.} CLIP, HQ-CLIP and BLIP-2 measures text-scene alignment (scaled by 100 for display). AvgObj indicates average object count per scene.}\label{quantitative_analysis}
\begin{tabularx}{\textwidth}{lYYYY}
\hline
Method & CLIP$\uparrow$ & HQ-CLIP$\uparrow$ & BLIP-2$\uparrow$ & AvgObj$\uparrow$ \\
\hline
GALA3D & 27.30 & 26.57 & 27.85 & 5.8 \\
\textbf{Ours} & \textbf{28.63} & \textbf{32.52} & \textbf{29.34} & \textbf{10.2} \\
\hline
\end{tabularx}
\end{table}

\section{Conclusion}

We present HiGS, a hierarchical scene generation framework based on multi-step associative semantic spatial composition, which progressively constructs 3D scenes from global layouts to local details with minimal user input. By introducing a progressive hierarchical spatial semantic graph (PHiSSG), HiGS organizes and propagates semantics throughout the generation process, ensuring global coherence, local controllability, and support for interactive, incremental editing. The PHiSSG maintains spatial and semantic consistency by explicitly modeling object dependencies and enforcing a one-to-one mapping between graph nodes and generated objects, thereby preserving geometric correctness across steps. Experimental results show that HiGS outperforms existing single-stage methods in layout rationality, style consistency, and alignment with user preferences, while significantly reducing the textual input required for comparable scene complexity.

\subsubsection{Limitation.}While HiGS demonstrates strong performance in multi-step generation and interactive editing, one key limitation remains. The semantic anchor used for region expansion may geometrically deviate from its counterpart in the global scene, which can lead to misalignment during integration.

\subsubsection{Future Work.}To address geometric inconsistencies between semantic anchors and global objects, future work will integrate geometric consistency constraints with instance-level alignment strategies to match shape, scale, and structure during generation, enabling more precise placement. 

\subsubsection{Acknowledgements.}Support and assistance from the Digital ART Lab at Shanghai Jiao Tong University during this research are gratefully acknowledged.

\newpage

%
%

\bibliographystyle{splncs04}
\bibliography{references}

\end{document}